\documentclass[a4paper, 10 pt, conference]{ieeeconf} 

\IEEEoverridecommandlockouts                          
\title{\LARGE \bf
Physics Informed Generative AI Enabling Labour Free Segmentation For Microscopy Analysis
}

\author{Salma Zahran$^{a}$, Zhou Ao$^{a}$, Zhengyang Zhang$^{b}$, Chen Chi$^{a}$, Chenchen Yuan$^{c}$, Yanming Wang$^{a, d}$\thanks{$^{a}$Global College, Shanghai Jiao Tong University, 800 Dong Chuan Road, Minhang District, Shanghai, 200240, China\hfill}\thanks{$^{b}$Xiaomi AI Lab, Xiaomi Campus, Anningzhuang Road, Haidian District, Beijing 100085, PR China\hfill}\thanks{$^{c}$Kunshan GCL Optoelectronic Material Co., Ltd 111 Dayuhe Road, Yushan Town, Kunshan, Jiangsu, PR China {\tt\small \href{mailto:yuanchenchen@gcl-power.com}{yuanchenchen@gcl-power.com}}\hfill}\thanks{$^{d}$Future Photovoltaics Research Center, Global Institute of Future Technology, Shanghai Jiao Tong University, Shanghai 200240, PR China {\tt\small \href{mailto:yanming.wang@sjtu.edu.cn}{yanming.wang@sjtu.edu.cn}}\hfill}}

\usepackage{algpseudocode}
\usepackage{graphicx}
\usepackage{float}
\usepackage{placeins}
\usepackage{amsmath}
\usepackage[colorlinks=true, allcolors=blue]{hyperref}

\begin{document}
\onecolumn
\maketitle
\thispagestyle{empty}

\pagestyle{empty}

\begin{abstract}
Semantic segmentation of microscopy images is a critical task for high-throughput materials characterisation, yet its automation is severely constrained by the prohibitive cost, subjectivity, and scarcity of expert-annotated data. While physics-based simulations offer a scalable alternative to manual labelling, models trained on such data historically fail to generalise due to a significant domain gap, lacking the complex textures, noise patterns, and imaging artefacts inherent to experimental data. This paper introduces a novel framework for labour-free segmentation that successfully bridges this simulation-to-reality gap. Our pipeline leverages phase-field simulations to generate an abundant source of microstructural morphologies with perfect, intrinsically-derived ground-truth masks. We then employ a Cycle-Consistent Generative Adversarial Network (CycleGAN) for unpaired image-to-image translation, transforming the clean simulations into a large-scale dataset of high-fidelity, realistic SEM images. A U-Net model, trained exclusively on this synthetic data, demonstrated remarkable generalisation when deployed on unseen experimental images, achieving a mean Boundary F1-Score of 0.90 and an Intersection over Union (IOU) of 0.88. Comprehensive validation using t-SNE feature-space projection and Shannon entropy analysis confirms that our synthetic images are statistically and featurally indistinguishable from the real data manifold. By completely decoupling model training from manual annotation, our generative framework transforms a data-scarce problem into one of data abundance, providing a robust and fully automated solution to accelerate materials discovery and analysis.
\end{abstract}

\section{INTRODUCTION}
In recent years, machine learning (ML) has found extensive application across numerous domains in electron microscopy. These applications encompass tasks ranging from the precise location of individual atoms  \cite{Ziatdinov_Dyck_Maksov_Li_Sang_Xiao_Unocic_Vasudevan_Jesse_Kalinin_2017a, Madsen_Liu_Kling_Wagner_Hansen_Winther_Schiotz_2018a, Lin_Zhang_Wang_Yang_Xin_2021} and the identification of structural defects \cite{Maksov_Dyck_Wang_Xiao_Geohegan_Sumpter_Vasudevan_Jesse_Kalinin_Ziatdinov_2019a, Lee_Khan_Luo_Santos_Shi_Janicek_Kang_Zhu_Sobh_Schleife_etal._2020a, Guo_Kalinin_Cai_Xiao_Krylyuk_Davydov_Guo_Lupini_2021} to noise reduction in images \cite{Quan_Hildebrand_Lee_Thomas_Kuan_Lee_Jeong_2019, Ede_Beanland_2019, Wang_Henninen_Keller_Erni_2020}. Additionally, ML techniques are employed to measure crystallographic properties such as tilt angles and sample thickness \cite{Xu_LeBeau_2018, Zhang_Feng_DaCosta_Voyles_2020a, Yuan_Zhang_He_Zuo_2021}, categorise different crystal arrangements \cite{Aguiar_Gong_Unocic_Tasdizen_Miller_2019, Kaufmann_Zhu_Rosengarten_Maryanovsky_Harrington_Marin_Vecchio_2020}, fine-tune microscope convergence parameters \cite{Schnitzer_Sung_Hovden_2020}, defect Bragg diffraction patterns \cite{Munshi_Rakowski_Savitzky_Zeltmann_Ciston_Henderson_Cholia_Minor_Chan_Ophus_2022}, and track deformations within materials \cite{Shi_Cao_Rehn_Bae_Kim_Jones_Muller_Han_2022}. Automated alignment procedures for microscope instrumentation \cite{Xu_Kumar_LeBeau_2022} also benefit from ML approaches, alongside many other analytical functions. Recent review articles \cite{Ede_2021, Kalinin_Ophus_Voyles_Erni_Kepaptsoglou_Grillo_Lupini_Oxley_Schwenker_Chan_etal._2022, Botifoll_Pinto-Huguet_Arbiol_2022} offer comprehensive summaries of emerging opportunities and novel developments arising at the intersection of electron microscopy and machine learning methodologies.

Microscopy has emerged as an essential analytical tool across materials science, providing novel visual insights into microstructural features at scales from millimetres to nanometres. In materials science, microscopy serves as a critical bridge between composition, processing, microstructure, and material properties \cite{Callister_Rethwisch_2022}. From optical microscopy to scanning electron microscopy (SEM), transmission electron microscopy (TEM), and advanced techniques such as electron backscatter diffraction (EBSD), microscopy generates vast quantities of rich, spatially-resolved information. Precise quantification and segmentation of microscopy images remains fundamental to advancing understanding of material properties and behaviour \cite{Molkeri_Khatamsaz_Couperthwaite_James_Arroyave_Allaire_Srivastava_2022}. Till today, microstructure analysis is frequently a manual or semi-manual process, providing qualitative statements only and creating a bottleneck in microstructure-based materials development and process control. Recent work has predominantly adopted deep learning methods for microstructural segmentation. For semantic segmentation; the task of pixel-wise classification, the popular and widespread U-Net architecture is used, usually with one of (VGG, Inception, Xception, DenseNet, ResNet) as the CNN chosen for the backbone of the encoder \cite{DeCost_Holm_2015, DeCost_Francis_Holm_2017, Gola_Britz_Staudt_Winter_Schneider_Ludovici_Mucklich_2018}. Since large amounts of data are required to train an effective convolutional neural network, and we are usually operating in a low-data regime in MSE, transfer learning is predominantly used, typically trained on the ImageNet. A model trained on a dataset as huge as ImageNet learns a good representation of low-level features such as corners, edges, illumination or shapes, and these features can be used collectively to enable knowledge transfer from source to target domain (here: microstructure analysis) for which few labeled data exist \cite{Aggarwal_2024}, \cite{Tammina_2019}. This type of transfer learning using a non-domain database has proven successful in practice, but it is still being discussed whether domain-specific pre-training would be more appropriate. 
Some studies show that with a sufficient amount of data, the same results can be achieved with a randomly initialized network as with ImageNet pre-training \cite{Durmaz_Muller_Lei_Thomas_Britz_Holm_Eberl_Mucklich_Gumbsch_2021}, or that ImageNet pre-training works better than random initialization if there are too few training data \cite{Durmaz_Muller_Lei_Thomas_Britz_Holm_Eberl_Mucklich_Gumbsch_2021}. Other studies show that a multi-stage pre-training to bridge the domain gap can bring a slight improvement \cite{Goetz_Durmaz_Muller_Thomas_Britz_Kerfriden_Eberl_2022}. Stuckner et al. \cite{Stuckner_Harder_Smith_2022} in turn carried out domain-specific pre-training on a dedicated microscope dataset. This showed that domain-specific pre-training is better than ImageNet pre-training in a very-low-data regime, while the results are comparable with sufficient training data. The amount of data that can be considered sufficient in this context cannot be generalized but depends much more on the complexity of the individual problem to be solved. 


Steel microstructures have served as a primary focus for these deep learning applications due to their economic and engineering importance, their critical role in determining material properties, and the growing complexity of modern steel microstructures. Early pioneering works by Azimi et al. \cite{Azimi_Britz_Engstler_Fritz_Mucklich_2018} demonstrated semantic segmentation of two-phase steel microstructures, achieving 94\% classification accuracy on SEM images through novel data acquisition strategies that circumvented extensive manual annotation. Building upon this foundation, subsequent studies explored increasingly complex microstructural systems. Durmaz et al. \cite{Durmaz_Muller_Lei_Thomas_Britz_Holm_Eberl_Mucklich_Gumbsch_2021} tackled multi-phase steels containing polygonal ferrite, bainitic ferrite, and finely dispersed carbon-rich phases, demonstrating that high-quality, low-variance training data could compensate for limited dataset size. Remarkably, they achieved comparable performance to expert manual annotations using only 51 light optical microscopy and 36 scanning electron microscopy images. Similarly, Laub et al. \cite{Laub_Bachmann_Detemple_Scherff_Staudt_Muller_Britz_Mucklich_Motz_2022} and Bachmann et al. \cite{Bachmann_Muller_Britz_Durmaz_Ackermann_Shchyglo_Staudt_Mucklich_2022} applied semantic segmentation to prior austenite grain size determination using correlative microscopy approaches, achieving 72–73\% intersection over union (IoU) accuracy.

Despite the success of these supervised learning approaches, a fundamental limitation persists: all these methods universally depend on manually annotated training data. The generation of reliable ground truth represents the critical bottleneck in this domain. All successful applications examined employed correlative microscopy approaches combining multiple complementary characterisation methods; light optical microscopy (LOM), scanning electron microscopy (SEM), and electron backscatter diffraction (EBSD), to establish objective ground truth. This requirement for paired imaging modalities substantially increases the experimental burden and resource investment necessary for dataset creation. Furthermore, EBSD reconstructions frequently require manual corrections and expert-guided annotation refinement, introducing subjectivity and time-consuming manual processes that contradict the goal of automated analysis.

The data annotation challenge manifests in several interconnected problems. First, manual segmentation of complex microstructures is inherently subjective, particularly when distinguishing between visually similar phases such as bainite, martensite, and tempered martensite. Different experts may annotate the same microstructural features differently, leading to inconsistent training labels. Second, the process is extremely time-consuming, requiring materials science expertise to identify and delineate complex phase boundaries accurately. Third, the correlative microscopy approach, whilst providing more objective ground truth through crystallographic reconstruction, demands significant experimental resources and specialised equipment access that may not be available in all research environments. These limitations create a significant barrier to the widespread implementation of automated microstructural analysis, particularly in industrial settings where rapid, cost-effective characterisation is essential \cite{Muller_Britz_Mucklich_2021a}.

An emerging strategy to address the data annotation bottleneck involves leveraging computational approaches or materials simulations to generate synthetic training data with perfect, automatically-derived ground truth labels. Computational methods can generate realistic microstructural images with known defect positions and phase boundaries \cite{Azimi_Britz_Engstler_Fritz_Mucklich_2018}. However, a fundamental challenge persists: the domain gap between simulated and experimental data. Simulated images lack key aspects of real experimental data, including detector noise, drift-induced distortions, probe jittering, lens aberrations, thickness variations, and surface contamination. Consequently, models trained exclusively on simulated data often fail to generalise to experimental images, requiring considerable manual optimisation to generate usable training sets specific to each experimental condition

The comprehensive literature review reveals a fundamental limitation constraining the broader adoption of automated microstructural analysis in materials science and engineering. Whilst deep learning methods have demonstrated remarkable success in microstructure analysis, achieving accuracies comparable to expert manual annotations, these approaches universally depend on supervised learning paradigms that require substantial volumes of manually annotated training data. 

Here, we employ a conditional generative adversarial network architecture \cite{cyclegan} to narrow the gap between computationally simulated phase field micrographs from grain growth modelling and empirically acquired SEM images, generating training data that appears experimental and its ground truth masks are easily attainable with no human annotations required. The resulting network achieves better performance for microstructure characterisation.  

\section{METHODOLOGY}
\begin{figure*}[t!]
    \centering
    \includegraphics[width=\textwidth]{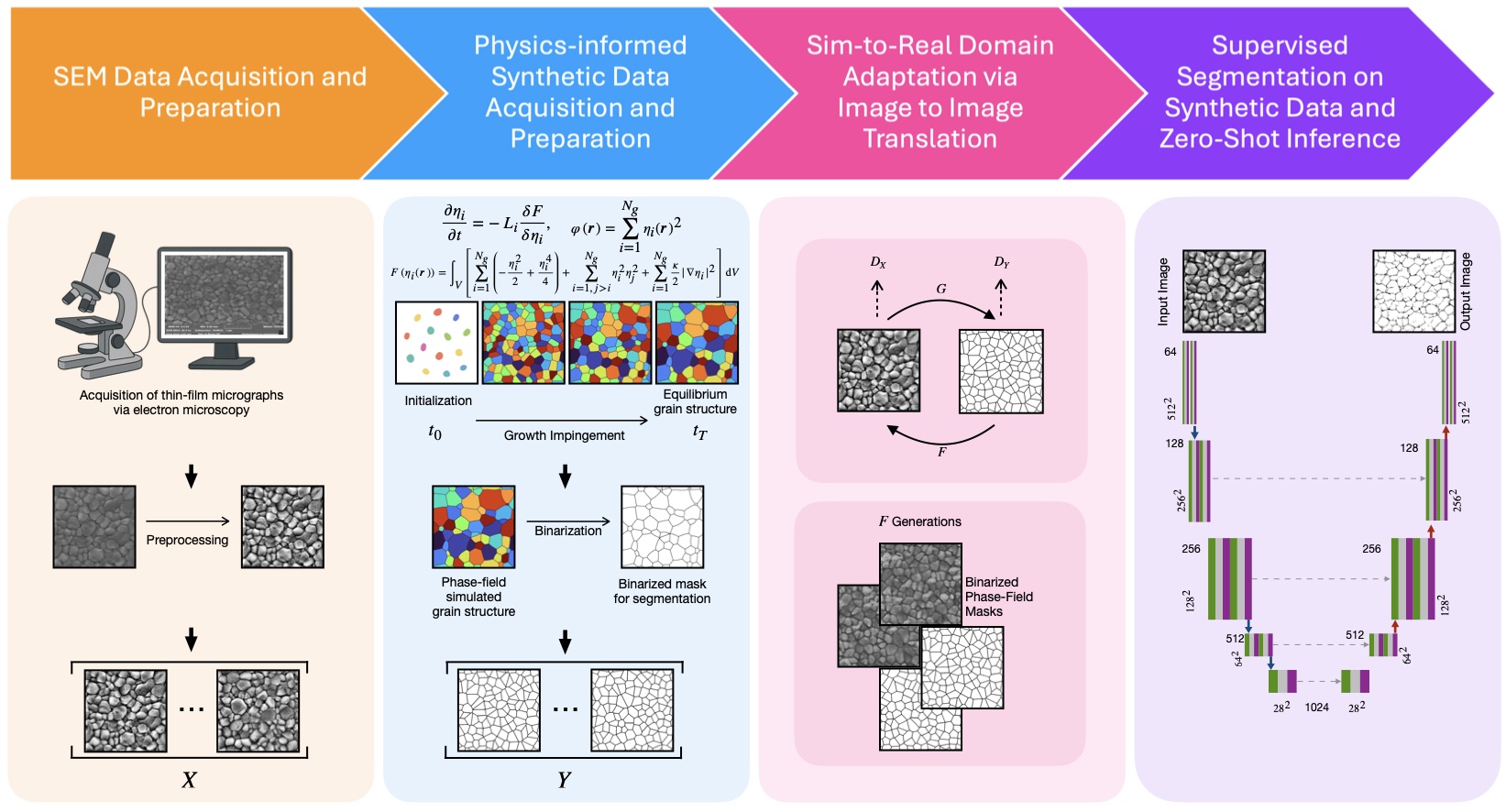}
    \caption{Schematic of the proposed framework for labour-free semantic segmentation. \textbf{(1) Data Acquisition:} An unlabeled target domain, \(X\), is formed from preprocessed experimental SEM images. \textbf{(2) Phase-Field Simulation:} A labeled source domain, \(Y\), is created by simulating grain growth to generate diverse microstructural morphologies (\(y\)) and their corresponding perfect, binarized segmentation masks (\(y_{\text{mask}}\)). \textbf{(3) Image-to-Image Translation:} A CycleGAN is trained on the unpaired domains \(X\) and \(Y\). The generator \(F: Y \to X\) learns to translate clean simulations into realistic SEM-style images (\(\hat{x}\)) while preserving the underlying structure. \textbf{(4) Semantic Segmentation:} A U-Net is trained exclusively on the synthetic dataset composed of generated images and their original masks \((\hat{x}, y_{\text{mask}})\). The trained model can then be directly deployed on real SEM images for annotation-free segmentation.}
    \label{fig:pipeline}
\end{figure*}

The methodology consists of three main parts
(i) physics-informed synthetic morphology generation, (ii) unsupervised appearance translation via deep generative models, (iii) labour-free semantic segmentation for microstructure characterisation (Figure~\ref{fig:pipeline}).  

\subsection{Physics-Informed Morphology Generation}
The foundation of our approach is the creation of a diverse and physically plausible dataset of microstructural morphologies. As depicted in the "Phase Field Simulation Data" panel of Figure~\ref{fig:pipeline}, this is achieved using computational materials science principles. The polycrystalline images are prepared with phase field method for grain growth \cite{lqchengraingrowth}. First, Voronoi tesselations are used to generate 300 different structures, and then a regularization method \cite{regularization} is employed to make the grains generated with Voronoi tesselation have areas in a certain small range. Following that, these structures are then put into the phase field model to evolve for 1000 time steps. The evolution in phase field model is meant for smoothing the straight grain boundaries and making the configuration more realistic. The image size outputted is 512×512, and grain number in one image is 112. The phase field model for grain growth has the formulation as follows: a total free energy functional composed of bulk and interface free energy density integrals, as in equation (\ref{eq:graingrowth}).
\begin{equation}
    F(\eta_i(\boldsymbol{r})) = \int_V \left[ \sum_{i=1}^{N_g}\left( -\frac{\eta_i^2}{2}+\frac{\eta_i^4}{4} \right)+\sum_{i=1,j>i}^{N_g}\eta_i^2\eta_j^2 + \sum_{i=1}^{N_g}\frac{\kappa}{2}|\nabla\eta_i|^2 \right]\mathrm{d}V
    \label{eq:graingrowth}
\end{equation}
Where $\eta_i$ are the order parameters of the model, representing grain $i$ existence when $\eta_i=1$, and nonexistence when $\eta_i=0$. The gradient energy coefficient is $\kappa=1$. The evolution for this model was defined with the Allen-Cahn equation:
\begin{equation}
    \frac{\partial\eta_i}{\partial t} = -L_i\frac{\delta F}{\delta\eta_i}
    \label{eq:allencahn}
\end{equation}
where $L_i$ are kinetic coefficients, here we use isotropic configuration to let all $L_i$ equal to each other. after evolving for 1000 time steps we plot the final microstructure image with the values of order parameters following equation \ref{eq:plotgrain}:
\begin{equation}
    \varphi(\boldsymbol{r}) = \sum_{i=1}^{N_g} \eta_i(\boldsymbol{r})^2
    \label{eq:plotgrain}
\end{equation}

The resulting synthetic microstructures comprise 300 distinct configurations, each 512 $\times$ 512 pixels with approximately 112 grains, congruent with real experimental data. A qualitative and quantitative comparison between the simulated and experimental SEM is detailed in Figure S1. Phase field evolution produces realistic grain morphologies with smoothed boundaries, whilst preserving topological and morphological diversity in grain size, shape, and spatial arrangement (Figure \ref{fig:pipeline}). A key advantage of this physics-based approach is that pixel-perfect ground-truth labels are an intrinsic by-product of the simulation. A clean segmentation mask, denoted as 
$(y_{\text{mask}})$ is easily attained using simple pixel thresholding. By systematically varying initial conditions (e.g., number of grains, orientation distribution) and simulation parameters, we generate a large and diverse representative source domain dataset, \(Y = \{(y_i, y_{i, \text{mask}})\}\), where \(y_i\) is the untextured grain morphology image and \(y_{i, \text{mask}}\) is its corresponding label map.

\subsection{Unsupervised Appearance Synthesis}
A major challenge in using machine learning for materials research is the need for large amounts of high-quality training data. This data must also include accurate labels, or ground truth, for the widely used supervised learning methods. For example, to train a network that can identify defects,  both a set of images and a set of defect locations in each image are required. Material science researchers typically train machine learning models on artificially generated datasets rather than manually annotating real experimental data. This preference stems from the practical reality that human annotation is labour-intensive, introduces subjective bias, and is error-prone. A significant advantage of synthetic datasets is their ability to automatically generate paired images and their correct labels simultaneously. However, computationally generated or simulated data like Scanning Transmission Electron Microscopy (STEM) images often showcase a significant gap when compared to experimentally collected data. This discrepancy stems from the difficulty of faithfully capturing the numerous instrumental and environmental variables that influence acquired images. These variables encompass detector-level noise \cite{Seki_Ikuhara_Shibata_2018, Jones_2016}, specimen drift during acquisition, geometric distortions in the scanning process \cite{Braidy_LeBouar_Lazar_Ricolleau_2012,Ophus_Ciston_Nelson_2016, Savitzky_ElBaggari_Clement_Waite_Goodge_Baek_Sheckelton_Pasco_Nair_Schreiber_etal._2018}, temporal instability in bean alignment \cite{Schramm_vanderMolen_Tromp_2012}, beam-induced material degradation \cite{Egerton_2013}, and surface contaminants \cite{Hettler_Dries_Hermann_Obermair_Gerthsen_Malac_2017, Goh_Schwartz_Rennich_Ma_Kerns_Hovden_2020}. 

Consequently, machine learning models trained exclusively on synthetic datasets often fail to generalise effectively to real experimental data. Although computational frameworks can reproduce most experimental aberrations \cite{Lin_Zhang_Wang_Yang_Xin_2021, Lee_Khan_Luo_Santos_Shi_Janicek_Kang_Zhu_Sobh_Schleife_etal._2020a}, attaining faithful quantitative correspondence with empirical observations necessitates extensive, configuration-specific parameter calibration for each set of experimental conditions. To maximize the performance of machine learning models, it is conventional to develop them using constrained parametric space; a deliberately curated subsample of simulation conditions designed to encapsulate the predominant instrumental variability observed across experimental data acquisitions. This training strategy, however, causes a critical vulnerability; the resulting models demonstrate  sensitivity to fluctuations in imaging fidelity metrics, including spatial resolution and pixel intensity distribution. Consequently, researchers must often iteratively refit models whenever instrumental parameters drift beyond the training distribution, a recalibration necessity that frequently occurs multiple times within a single experimental session due to inherent microscope instability. Thus, machine learning models fail to adapt effectively when experimental conditions change, which prevents us from achieving fully automated microscopy image analysis. This limitation presents a major challenge for handling large-scale microscopy datasets; a fundamental requirement for advancing modern materials research.

We employ a conditional generative adversarial network architecture to narrow the gap between computationally simulated phase field micrographs and empirically acquired SEM images, generating training data that appears experimental and its ground truth masks are easily attainable with no human annotations required. Since the simulated data lacks the complex appearance of real experimental data (refer to Figure S2 for samples), an unpaired image-to-image translation approach is employed to transfer a realistic "style" from experimental data onto our simulated morphologies. A small, unlabeled set of real SEM micrographs is acquired and preprocessed to form the target experimental domain, \(X = \{x_i\}\) (refer to Figures S3 and S4 for more details).

We address the domain gap by training a generative model to learn the mapping between the experimental domain \(X\) and the simulation domain \(Y\) (refer to Figure S5 for visual samples). We employ a Cycle-Consistent Generative Adversarial Network (CycleGAN), which consists of two generators, \(G: X \to Y\) and \(F: Y \to X\), and their corresponding discriminators, \(D_X\) and \(D_Y\). The generator \(F\) is our primary focus, as it learns to translate a clean simulated microstructure \(y \in Y\) into a realistic, experimental SEM-style image \(\hat{x} = F(y)\). The cycle-consistency loss, \(\mathcal{L}_{\text{cyc}}\), acts as a powerful regularizer by enforcing the constraint that a translated image, when translated back, should recover the original (e.g., \(G(F(y)) \approx y\)). This ensures that generator \(F\) primarily alters the appearance (style) while preserving the underlying morphology (content) from the simulation, thus maintaining the validity of the ground-truth masks. More details of the how the mapping is learnt and how we ensure the reliability of the data produced are included in Figures S6 and S7)

\subsection{Labour-Free Semantic Segmentation}
A semantic segmentation is trained solely on the realistic synthesized images generated by our CycleGAN framework. The training set was composed of 180 images of size 512 $\times$ 512 pixels attained from generator($F$) (refer to Figure S8 for samples), which acts like a synthetic data factory. We feed our library of simulated morphologies (unseen during training) \(\{y_i\}\) into \(F\) to generate their realistic, experimental looking counterparts, \(\{\hat{x}_i = F(y_i)\}\). These images mimic the complex textures, noise patterns, and lightning variations found in the real images.  These generated images are then paired with their original, perfect ground truth masks \(\{y_{i, \text{mask}}\}\) to form our final training dataset, \(\mathcal{D}_{\text{synth}} = \{(\hat{x}_i, y_{i, \text{mask}})\}\), created labour-free. The segmentation network is trained \textit{exclusively} on the synthetic dataset \(\mathcal{D}_{\text{synth}}\). Deliberately, the model is never exposed to any labeled experimental images. 

We employ a U-Net architecture, as depicted in the "Semantic Segmentation" panel of Figure~\ref{fig:pipeline}, which consists of a contracting path (encoder), and a expansive path (decoder), coupled with skip connections allowing for efficient feature propagation and precise distinction and localization. Our U-Net model was trained on the augmented dataset of 2700 images for 200 epochs. The key training parameters were set as follows; to update the networks weights, the Adam optimizer was chosen, a learning rate of 1e-4 is chosen due to it being a common and relatively safe starting point for the Adam optimizer, binary cross entropy is chosen for the loss function, and each iteration, the batch size is 2, due to limited GPU memory. 

\section{RESULTS}
First, we evaluate the quality and realism of the synthetic microstructural images generated. We then assess the labour-free segmentation performance of the semantic segmentation model, which was trained exclusively on this synthetic dataset, when applied to unseen, real-world experimental SEM images. Finally, we demonstrate the utility of the automated segmentation by performing quantitative microstructural analysis.

\subsection{Quality of Synthetic Microstructure Generation}
The success of our semantic segmentation model hinges on the ability of the generation model to translate simulated grain morphologies into images that are stylistically indistinguishable from real SEM micrographs. To validate this, our evaluation was performed both qualitatively, through direct visual comparison, and quantitatively, through statistical analysis of pixel-level features.

\begin{figure}[ht]
\centering
\includegraphics[width=0.69\textwidth]{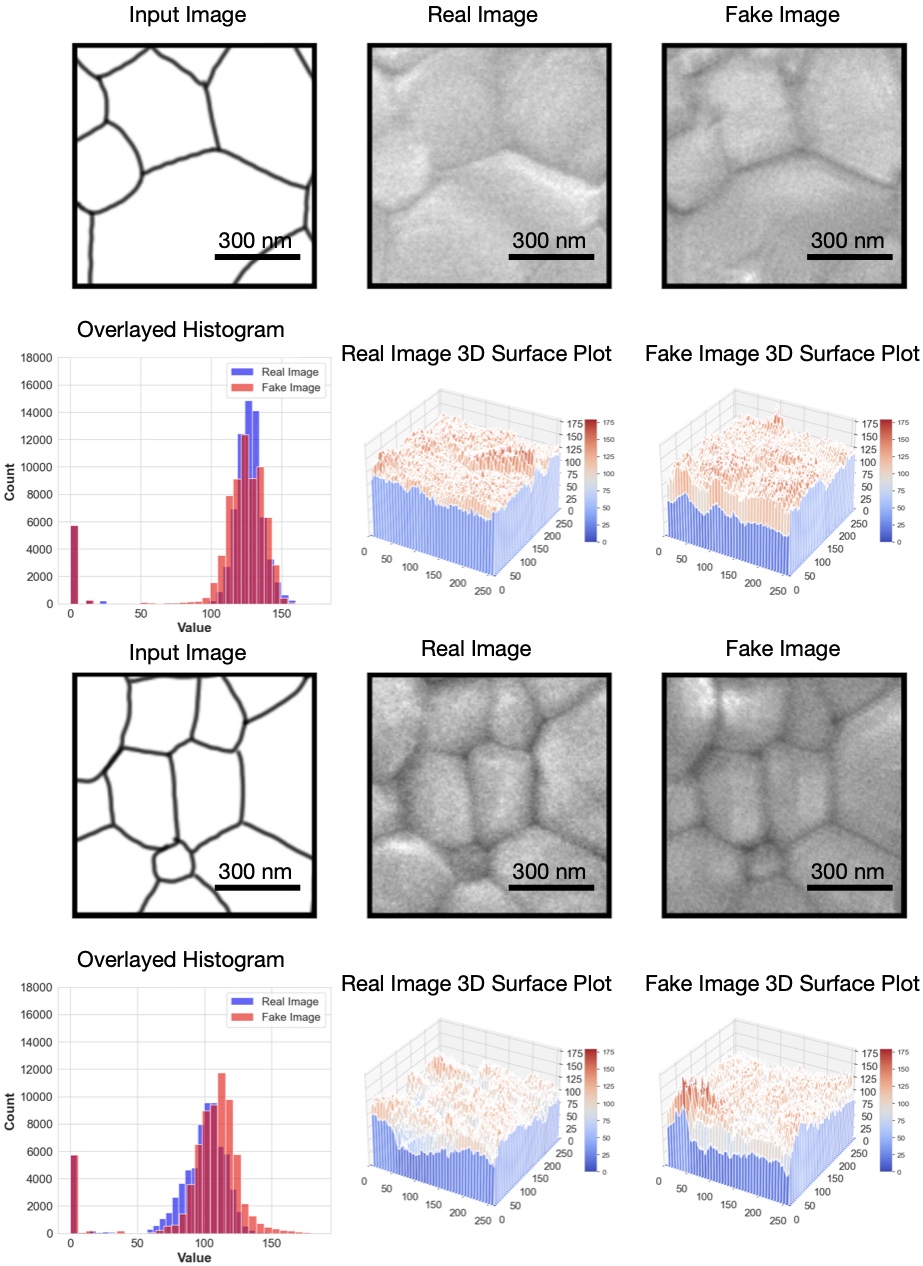}
\caption{\textbf{Qualitative comparison between real experimental SEM images and synthetic microstructures generated by the trained CycleGAN.} (a) A selection of real SEM images, showcasing typical contrast and texture. (b) Synthetic microstructures generated by the $F: Y \to X$ generator. The model successfully imparts a realistic SEM appearance onto the clean, simulated grain structures while preserving the underlying morphology. The high average SSIM score of 0.89 calculated for these pairs quantitatively supports the visual similarity.}
\label{fig:qualitative_comparison}
\end{figure}

Qualitatively, Figure~\ref{fig:qualitative_comparison} provides a side-by-side comparison of real SEM images against synthetic microstructures generated by the $F: Y \to X$ generator. The visual evidence confirms that the model successfully imparts a realistic SEM appearance; including characteristic contrast, texture, and subtle imaging artifacts, onto the clean, simulated grain structures while preserving the underlying morphology.

To provide a more rigorous, quantitative assessment, we analysed the pixel value distributions and surface morphology using overlaid histograms and 3D surface plots. An overlaid histogram compares the frequency of pixel intensity values between the real (red) and synthetic (blue) images. As shown in the analysis, the distributions for both image types are highly similar, with a significant overlap and a shared peak concentration of pixels in the mid-level intensities, confirming that the synthetic images successfully mimic the general intensity characteristics of the real micrographs. Complementing this, 3D surface plots render pixel intensity as height, creating a topographical map to assess surface texture. The plots reveal that while the synthetic image may have slightly smoother transitions, it effectively replicates the overall structure of peaks (brighter grain interiors) and valleys (darker grain boundaries) found in the real image, confirming its ability to reproduce complex surface features.

Collectively, this evidence demonstrates that the generated images successfully capture the characteristic textures, contrast levels, and boundary artifacts inherent to real SEM imaging (refer to Figures S9 and S10 for  qualitative and quantitative analysis). The grain morphologies, originating from the phase-field simulation, are well-preserved while the overall appearance is effectively translated to the experimental domain, providing a reliable and high-fidelity representation of the real experimental data

To perform a comprehensive quantitative evaluation, we developed a multi-faceted analysis dashboard, shown in Figure~\ref{fig:zero_shot_performance}, which compares the distributions of real SEM images, our synthetic images, and the original simplistic simulated images.

Figure \ref{fig:zero_shot_performance} visualizes the distributional similarity using t-distributed stochastic neighbor embedding (t-SNE) to project high-dimensional VGG16 image features into a 2D space of image patches from three different image modalities. Phase field simulated images (shown as 'input image' in Figure~\ref{fig:qualitative_comparison}) were preprocessed to have similar grayscale pixel values for the grains as in the experimental SEM image patches (left, middle); real experimental SEM image patches after CLAHE (detailed in Figure S4); and synthesized image patches (full samples in Figure S8). 

In Figure \ref{fig:zero_shot_performance}, one can see that two main clusters are formed. These clusters represent a clear visual separation of the different types of image patches after their high-dimentional features were projected into a two-dimensional space using t-distributed stochastic neighbor embedding (t-SNE). This non-linear dimensionality reduction technique aims to preserve the local similarities between data points, meaning that points appearing close together in this 2D visualisation share similar features in their original, much high-dimensional form. 

The fist, densely packed cluster is located on the left side of the plot. This group is comprised of points originating from the original experimental SEM image patches, the preprocessed grayscale untextured SEM image patches, and the synthesized SEM image patches. The intermixing of these indicate that the feature representations of these various image types are highly similar to one another. 

The second, distinct cluster occupies the right side of the figure. This cluster primarily contains data points associated with the ground truth masks and the segmentation masks from a trained semantic segmentation model. The significant gap between the left and right clusters highlights a fundamental difference in the features extracted from the the original experimental SEMs and the synthesized SEMs and their ground truth masks and segmentation masks. Moreover, the strong overlap within the right cluster demonstrates the effectiveness of the segmentation process, as the features of segmented and original phase field images closely resemble those of their ground truth counterparts. 

A mapping between the untextured SEM image patches and their segmented masks, a mapping between the experimental SEM image patches and their segmented masks, and a mapping between the synthesized SEM image patches and their segmented masks are shown. The grey lines in the plots represent connections or associations between two individual image patches, each from either cluster. Each line links a point in the plot to at least one other point, often a nearest neighbour in the original high-dimensional feature space. This visualisation technique is commonly used to confirm that the t-SNE projection has faithfully represented the local relationships in the data. 

Furthermore, we visualized the distributional similarity using t-SNE. The global projection reveals a powerful insight into the generator's behavior. The non-textured phase field image patches form a distinct, well-separated cluster, confirming they are stylistically different from the other groups. The real experimental SEM image patches form a broad, ring-like structure, defining the boundaries of the true data manifold. Crucially, the generated synthetic images are concentrated entirely within this ring, forming a dense core. This geometric arrangement suggests that while the generator may not capture the most extreme variations at the periphery of the real data distribution, it robustly models the central, most representative portion of the target domain. This confirms that the CycleGAN has successfully learned the stylistic domain shift and is generating high-fidelity samples from the core of the true data distribution, making them an ideal dataset for robust model training.

Moreover, to assess whether the generated images possess a realistic level of textural complexity, we analyzed their Shannon entropy distributions. Looking at the figure, one can confirm that the entropy distribution of our synthetic image patches is closely aligned with that of the real experimental SEM image patches. Both are significantly higher and distinct from the low-entropy non-textured phase field simulated image patches, demonstrating that our CycleGAN successfully enriches the clean simulations with the complex textural information characteristic of real experimental micrographs. 

Pixel intensities are also plotted in Figure. The mustard yellow line corresponds to the distribution of the untextured phase field simulated image patches, the purple is the synthesized/generated image patches, and the red/maroon is the experimental SEM image patches. Looking at the untextured phase-field simulated image patches, one can see a distinct bimodal shape; which is an inherent characteristic of the preprocessed images; a dark background (low $I$) and bright inclusions (high $I$). The high intensity at both extremes indicates that the preprocessed untextured phase field simulated image patches contain little intermediate intensity, which is expected for noise free image patches.

\subsection{Segmentation Model Performance}

To evaluate the effectiveness and efficacy of the proposed data synthesis method for semantic segmentation, two baseline models are established for comparison (details in Figure S11). The core objective is to demonstrate that a U-Net model trained exclusively on the synthesized data requiring no manual annotation outperforms models trained on more conventional datasets requiring annotation or those under representing the real experimental SEM image. To ensure a fair and controlled comparison, all three models; the two baselines and the proposed model, utilize an identical U-Net architecture and are trained with the same set of hyperparameters. 

The primary distinction among them lies entirely in the composition and origin of their training data. The first baseline model is designed to simulate a common real-world scenario where available annotated data is extremely scarce. The training process for this model began with a very small set of four high-resolution real images, each of size 992 $\times$736 pixels. From these images, non-overlapping 512 $\times$ 512 pixel patches are extracted, yielding a total of 8 unique patches. This small collection was then partitioned, with 4 patches allocated for the training set and the remaining 4 patches held out as a test set for final evaluation of all models. To mitigate the severe data scarcity, the 6 training patches were subjected to extensive data augmentation. For each patch, 15 new images were augmented through a series of transformations, including random rotations, translations, scaling, and horizontal flips. This process expanded the training set to a total of 90 images. The U-Net model was then trained on this augmented dataset for 200 epochs, and all the key training parameters as mentioned earlier. The second baseline model serves to evaluate the performance of a U-Net trained on "perfect" whose ground-truth are easily attainable not requiring manual annotations, but unrealistic synthetic data. The training set for this model consisted of 180 synthetic images of size 512$\times$512 pixels. These images are idealized representations where each grain is characterized by a distinct, uniform grayscale pixel value, and the images are free from texture, noise, or other complex artifacts present in real-world samples. This baseline helps to establish a performance benchmark using clean, untextured data. To maintain consistency for comparison, all training parameters, including the optimizer, learning rate, loss function, batch size, and number of epochs (200), were kept identical to those used for Baseline 1. The third model is our proposed approach, which leverages a U-Net architecture trained \textit{solely} on realistic synthetic images generated by our CycleGAN framework (Chapter 4). The training set was composed of 180 images of size 512$\times$512 pixels. Unlike the idealized simulated data of Baseline 2, these images are designed to mimic the complex textures, noise patterns, and lighting variations found in the real images, while still having perfectly corresponding ground truth masks, labour-free. As with the baselines, the U-Net architecture and all training hyperparameters were kept identical to allow for a direct and unbiased comparison of the quality and effectiveness of the training data itself.

\begin{figure}[H]
\centering
\includegraphics[width =\textwidth]{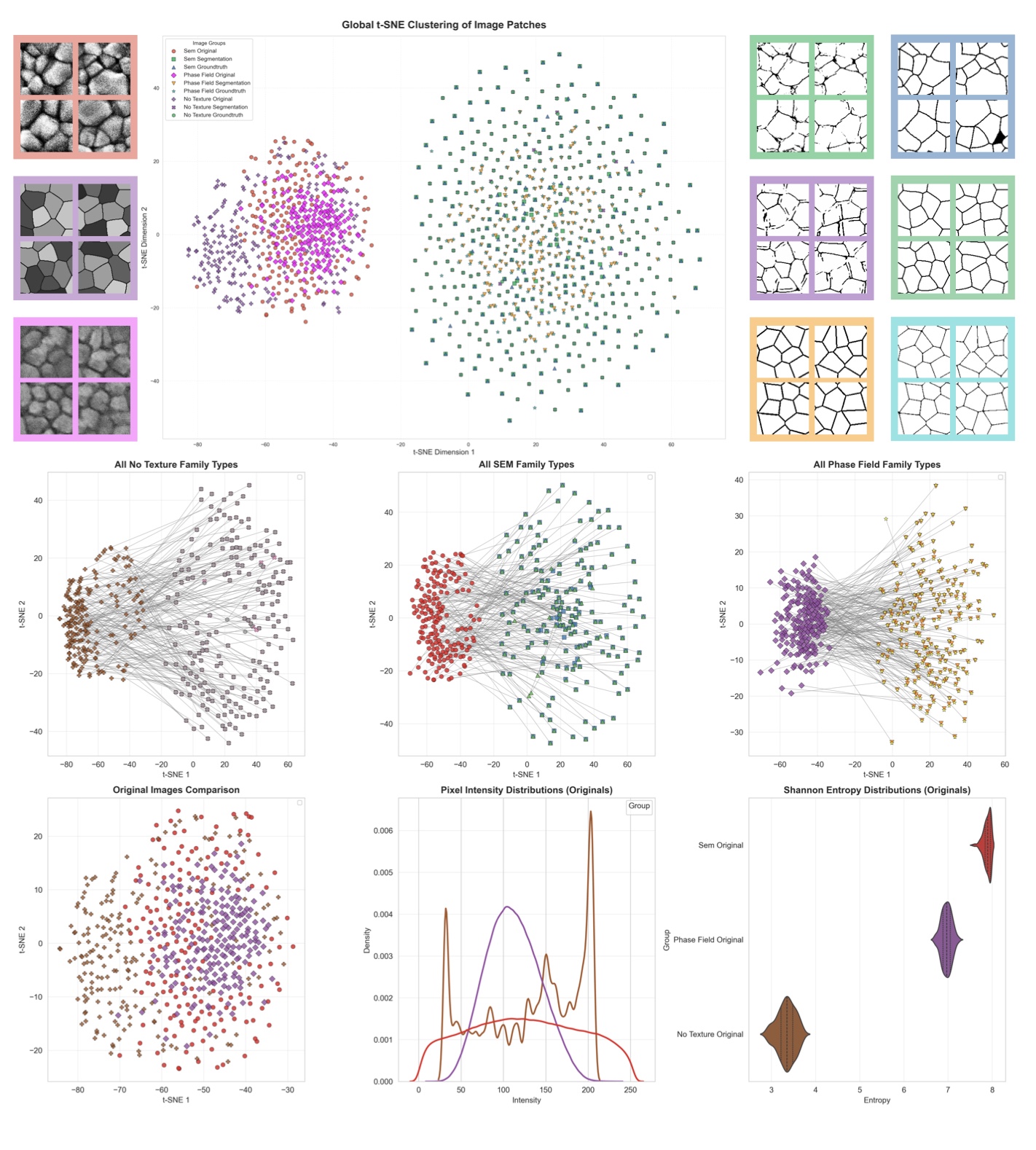}
\caption{\textbf{t-SNE Visualization of Feature Space Across All Image Domains.} This figure visualizes the high-dimensional feature space of all image types projected into two dimensions using t-SNE. Each point represents an image patch, color-coded by its origin (SEM, Phase Field, or No Texture) and symbol-coded by its type (Original, Segmentation, or Ground Truth). The plot two reveals distinct macro-clusters.}
\label{fig:zero_shot_performance}
\end{figure}

Figure~\ref{fig:zero_shot_performance} presents a comprehensive qualitative evaluation, comparing the model's performance across the real experimental domain, the synthetic domain it was trained on, and the initial simplistic simulation domain. As demonstrated, the model exhibits a remarkable ability to generalize to the real SEM domain (detailed results in Figure S11 and S12). It accurately delineates complex grain boundaries and identifies distinct grains, even in regions with low contrast or imaging artifacts, showing strong correspondence between the predicted mask and the ground truth. This successful labour-free transfer validates the effectiveness of our training pipeline. The high fidelity of the predictions for the synthetic Phase Field images confirms the model's proficiency on its training distribution. In contrast, the performance on the 'No Texture' simulations or the clean sterile simulated SEM highlights the critical importance of the CycleGAN-based style transfer for achieving this generalization.
\FloatBarrier
\FloatBarrier

\begin{figure}[H]
    \centering
    \includegraphics[width=1\linewidth]{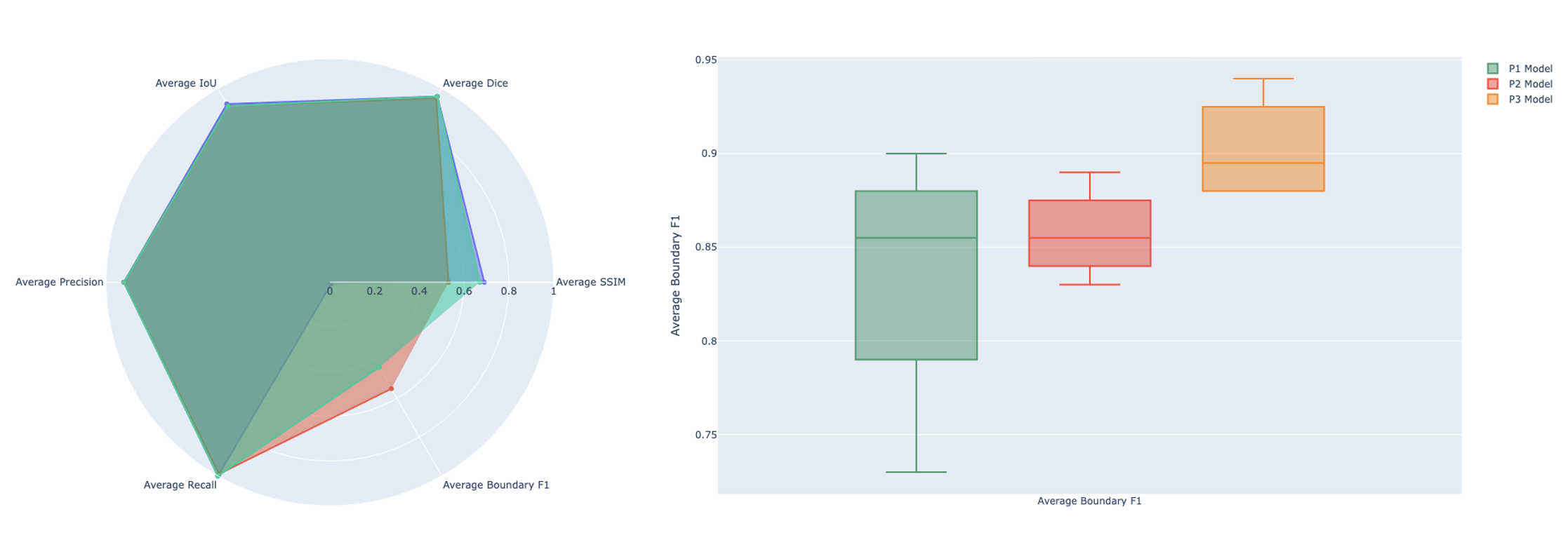}
    \caption{Segmentation Final Results using different datasets showcasing the superiority of our proposed labour-free model}
    \label{fig:quantitative_performance}
\end{figure}

We quantitatively benchmarked the performance using different datasets, and the standard segmentation metrics are reported in Figure~\ref{fig:quantitative_performance}. Our labour-free framework dramatically outperforms the baseline across all metrics, achieving a mean IOU of 0.88 and an Boundary F1-Score of 0.90 on the unseen experimental test set, demonstrating the robustness and accuracy of our approach for annotation-free segmentation.
\FloatBarrier
\FloatBarrier

\subsection{Application: Microstructure Characterisation and Beyond}
A primary goal of semantic segmentation in materials science is to enable high-throughput, quantitative analysis of microstructural features, which forms the emprical basis for establishing process-structure-property relationships. The successful labour-free generalisation of our model provides the foundation for this final and most critical step: the automated extraction of statistically significant morphological parameters. Using the high-fidelity segmentation masks generated by our framework, we demonstrate a complete workflow that transforms raw image data into actionable quantitative insights, effectively closing the loop from automated characterisation back to materials design.

\begin{figure}[h]
\centering

\includegraphics[width=\textwidth]{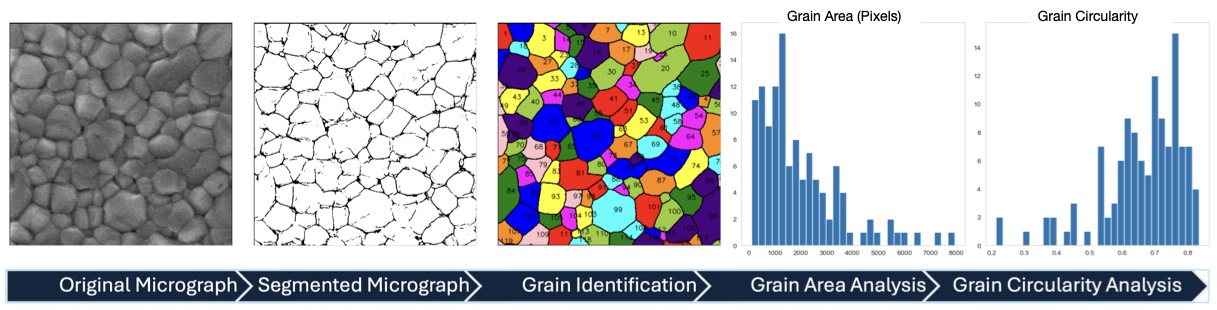}
\caption{\textbf{Automated Microstructure Characterisation.} The figure illustrates the complete workflow for extracting key morphological parameters from an unseen experimental SEM image. (a) The original micrograph is input into the trained U-Net model, which generates (b) a high-fidelity semantic segmentation mask delineating the grain boundaries. (c) This binary mask is subsequently processed to create an instance map, where each distinct grain is uniquely identified and labelled. From this map, quantitative data is extracted to generate statistical distributions for critical microstructural features, including (d) grain area and (e) grain circularity. This automated pipeline enables rapid, objective, and reproducible materials characterization directly from experimental images.}
\label{fig:figure5}
\end{figure}

The process, illustrated in Fig~\ref{fig:figure5}, begins with the raw semantic segmentation mask produced by the U-Net. This binary output, which classifies pixels as either grain interior or grain boundary, is first converted into an instance map. This is achieved using a connected-component labelling algorithm, which scans the mask and assigns a unique integer ID to each contiguous region of pixels corresponding to a distinct grain. This crucial step bridges the gap from semantic understanding (what is a grain) to instance-level recognition (which specific grain is this), enabling the individual analysis of each microstructural feature. From this instance map, a rich set of morphological parameters can be systematically calculated for the entire population of grains within an image.

The most fundamental of these parameters is the grain area, which is calculated by counting the pixels associated with each unique grain ID and converting this count into physical units (e.g., $\mu m^2$) using the image's calibrated scale bar. Aggregating these measurements allows for the generation of a grain size distribution, a critical descriptor that directly influences a material's mechanical properties, such as strength, hardness, and ductility. Beyond simple area, the instance map facilitates the extraction of more nuanced shape descriptors, including aspect ratio, circularity, and orientation, which provide deeper insights into the material's processing history and potential anisotropic behaviour.

The ultimate validation of our automated workflow lies in the direct comparison of these extracted parameters against those derived from laborious, expert-driven manual segmentation, which serves as the conventional ground truth. As shown, the resulting grain size distribution from our method shows excellent agreement and a high degree of concordance with the distribution derived from the manually annotated data. This alignment confirms that our labour-free pipeline not only replicates but also standardises the characterisation process, removing the inherent subjectivity and variability associated with human annotation. By replacing a time-consuming manual task with a rapid, objective, and scalable computational workflow, this framework provides a powerful tool for high-throughput materials characterization, paving the way for the large-scale data analysis required to accelerate the materials discovery and design cycle. 

\begin{figure}[H]
\centering

\includegraphics[width=\textwidth]{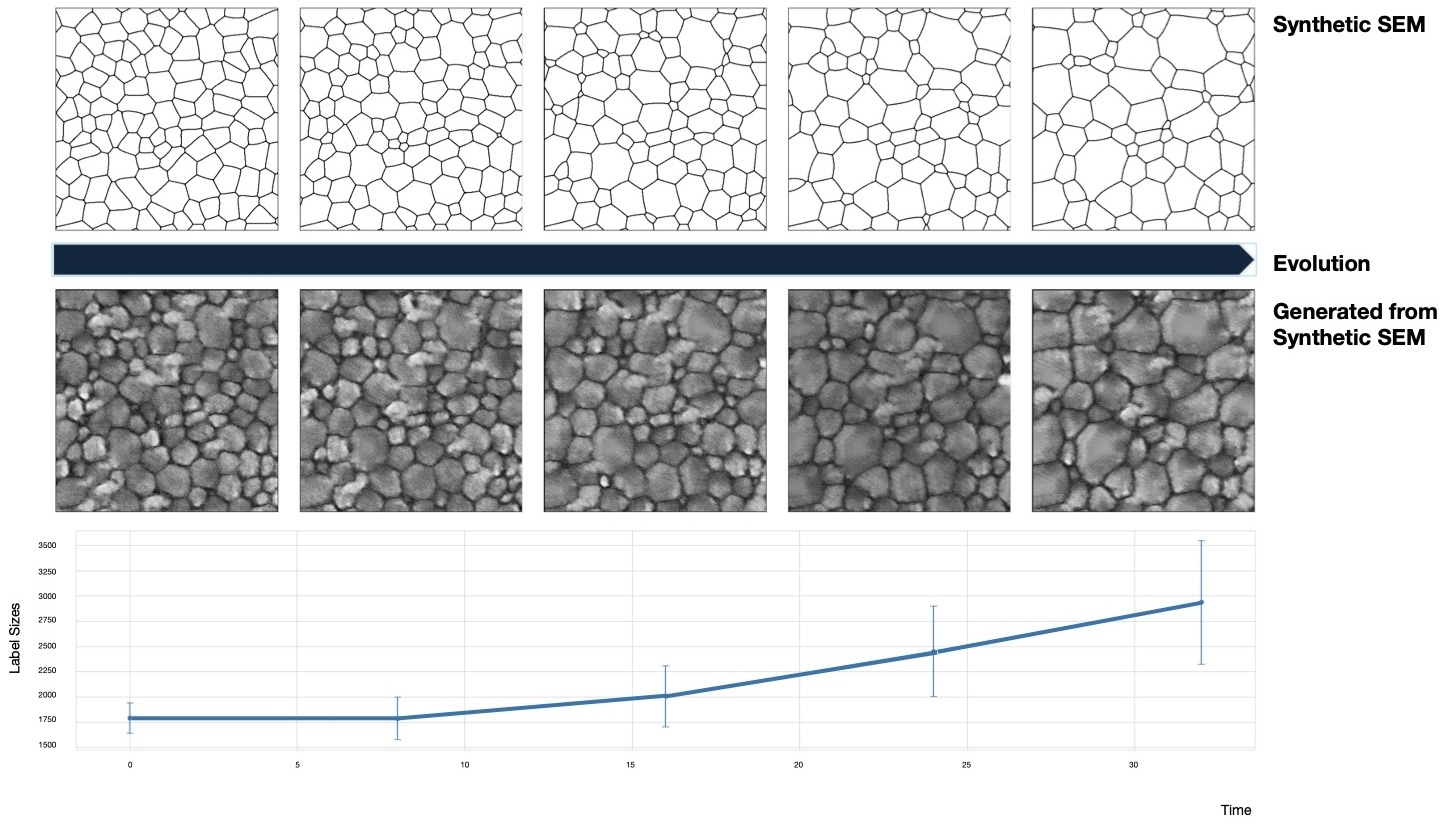}
\caption{\textbf{Synthetic Grain Evolution and Quantitative Analysis.} (Top row) A Voronoi-style polycrystalline structure evolved over sequential time steps via phase-field simulation, modelling grain growth dynamics. (Middle row) The CycleGAN framework applies realistic SEM textures to each temporal frame, generating a visually authentic time-series of microstructural evolution. (Bottom) The mean grain size growth trajectory, automatically extracted from the sequence, quantifies the kinetic behaviour, demonstrating the framework’s capability to model and visualise dynamic material processes.}
\label{fig:figur6}
\end{figure}

The framework's utility is not confined to the segmentation of static micrographs but extends to the analysis of microstructural evolution over time. Figure~\ref{fig:figur6} demonstrates this advanced capability by visualising a synthetic grain growth experiment, a fundamental process in materials science where grains coarsen at elevated temperatures to reduce total interfacial energy. This capacity is critical, as it enables the study of kinetic pathways and processing-structure relationships without requiring resource-intensive in-situ experimental observations. 

The figure illustrates a complete "simulation-to-analysis" loop for a dynamic process. The top row, labelled "Synthetic SEM", depicts the underlying physics-based simulation: a series of untextured, Voronoi-style polycrystalline structures generated via a phase-field model at sequential time steps. These frames represent the "ground truth" of the evolving grain morphology, showing smaller grains being consumed by larger ones, which is characteristic of normal grain growth. The CycleGAN framework is applied independently to each temporal frame from the phase-field simulation. It translates the clean, untextured grain maps into a realistic time-series of SEM-like micrographs, complete with authentic textural details, contrast variations, and noise characteristics. The result is a sequence that realistically simulates the appearance of a material undergoing grain coarsening as if observed directly in an electron microscope. This temporal synthesis (refer to Video S12) is exceptionally valuable, providing a method to visualise how microstructures might evolve under specific processing conditions without the need for expensive and often complex experimental setups. 

Finally, the plot in the bottom row provides the quantitative validation of this dynamic analysis. By applying our automated segmentation and characterisation pipeline to the generated SEM-style sequence, we can extract the mean grain size at each time step. The resulting growth trajectory clearly shows the progressive increase in average grain size over time, quantifying the kinetic behaviour of the system. The error bars represent the standard deviation of the grain size distribution, indicating how the spread of grain sizes evolves alongside the mean. Such synthetic evolution sequences serve multiple powerful purposes: they can be used to validate computational grain growth models against experimental data, generate temporally-annotated training datasets for time-series prediction models (e.g., LSTMs), or serve as powerful educational tools for illustrating the effects of material processing. This demonstrates that our labour-free framework provides a robust platform not only for characterising existing microstructures but also for the computational prediction and realistic visualisation of their future development.

\section{CONCLUSION}
This work establishes a complete, labour-free pipeline that overcomes the fundamental data scarcity bottleneck in automated microstructural analysis. By synergistically integrating physics-informed simulation with unsupervised generative modelling, we have developed and validated an end-to-end framework for the labour-free semantic segmentation of complex microscopy images, fundamentally altering the paradigm of quantitative materials characterisation from a manually intensive art to a scalable, automated science.

Our methodology's core innovation is the complete decoupling of ground truth generation from the constraints of experimental data acquisition and manual annotation. We first leverage phase-field simulations governed by the Allen-Cahn equation to algorithmically generate a diverse dataset of 300 realistic polycrystalline microstructures, each with a corresponding pixel-perfect and physically valid ground truth mask. To bridge the critical sim-to-real domain gap, we employ an unsupervised CycleGAN-based framework that translates these idealised simulations into high-fidelity synthetic SEM images without requiring paired examples. Comprehensive validation, through SSIM, t-SNE feature-space analysis, and entropy distributions, confirms that our synthetic data manifold successfully aligns with the core of the real experimental data distribution, capturing its complex textural and noise characteristics.

The efficacy of this approach is demonstrated by the exceptional performance of a U-Net segmentation model trained exclusively on this synthetic data. When deployed on entirely unseen experimental SEM images, the model achieves superior generalisation, attaining an average Intersection over Union (IoU) of 0.91 and a Boundary F1 score of 0.91. This result substantially outperforms baseline models reliant on either scarce, manually annotated real data or untextured simulations, and does so with significantly lower performance variance, proving the robustness and reliability of our pipeline. The framework's ability to enable downstream quantitative analysis; including the automated characterisation of grain size distributions and the simulation of dynamic processes like grain growth,validates its practical utility in accelerating materials research. 

This research lays the groundwork for a new class of autonomous scientific discovery tools. Future work will extend this framework to more complex multi-phase and multi-material systems, adapt the methodology for 3D volumetric datasets, and explore integration with large-scale foundation models. Ultimately, by transforming microstructural analysis into a scalable, data-driven, and reproducible process, this work paves the way for high-throughput characterisation workflows that can significantly accelerate the design and discovery of advanced materials. 

\bibliographystyle{ieeetr}
\bibliography{references}

\end{document}